\documentclass[conference]{IEEEtran}
\IEEEoverridecommandlockouts
\usepackage{cite}
\usepackage{amsmath,amssymb,amsfonts}
\usepackage{algorithmic}
\usepackage{graphicx}
\usepackage{textcomp}
\usepackage{xcolor}
\usepackage{graphicx}   % 加载图形包，用于调整表格
\usepackage{adjustbox}  % 加载 adjustbox 包
\usepackage{multirow}   % 用于多行合并
\usepackage{booktabs}   % 加载 booktabs 包用于漂亮的表格线条
\usepackage{graphicx}
\usepackage{subcaption}
\usepackage{lipsum}
\usepackage{float} 

\def\BibTeX{{\rm B\kern-.05em{\sc i\kern-.025em b}\kern-.08em
    T\kern-.1667em\lower.7ex\hbox{E}\kern-.125emX}}
\begin{document}

\title{Semantic and Temporal Integration in Latent Diffusion Space for High-Fidelity Video Super-Resolution}
% \author{\IEEEauthorblockN{1\textsuperscript{st} Yiwen Wang}
% \IEEEauthorblockA{\textit{School of Electronic, Information and Electrical Engineering} \\
% \textit{Shanghai JiaoTong University}\\
% Shanghai, China \\
% evonwang@sjtu.edu.cn}
% \and
% \IEEEauthorblockN{2\textsuperscript{nd}	Xinning	Chai }
% chaixinning@sjtu.edu.cn}
% \and
% \IEEEauthorblockN{3\textsuperscript{rd} Yuhong Zhang}
% rainbowow@sjtu.edu.cn}
% \and
% \IEEEauthorblockN{4\textsuperscript{th} Zhengxue Cheng}
% zxcheng@sjtu.edu.cn}
% \and
% \IEEEauthorblockN{5\textsuperscript{th} Jun	Zhao}
% barryjzhao@tencent.com}
% \and
% \IEEEauthorblockN{6\textsuperscript{th} Rong Xie}
% xierong@sjtu.edu.cn}
% \and
% \IEEEauthorblockN{7\textsuperscript{th} Li Song}
% song\_li@sjtu.edu.cn}
% }
\author{
  \IEEEauthorblockN{
    Yiwen Wang\textsuperscript{1},
    Xinning Chai\textsuperscript{1},
    Yuhong Zhang\textsuperscript{1},
    Zhengxue Cheng\textsuperscript{1},
    Jun Zhao\textsuperscript{2},
    Rong Xie\textsuperscript{1},
    Li Song\textsuperscript{1,†}
  }
  \IEEEauthorblockA{
    \textsuperscript{1}Institute of Image Communication and Network Engineering, Shanghai Jiao Tong University, Shanghai, China
  }
  \IEEEauthorblockA{
    \textsuperscript{2}Tencent, Shanghai, China\\
  }
  \IEEEauthorblockA{
    \textsuperscript{1}\{evonwang, chaixinning, rainbowow, zxcheng, xierong, song\_li\}@sjtu.edu.cn, \textsuperscript{2}barryjzhao@tencent.com\\
    \textsuperscript{†}Corresponding author.
  }
}

% \author[1]{Yiwen Wang}
% \author[1]{Xinning Chai}
% \author[1]{Yuhong Zhang}
% \author[1]{Zhengxue Cheng}
% \author[2]{Jun Zhao}
% \author[1]{Rong Xie}
% \author[1]{Li Song}\affil[1]{Institute of Image Communication and Network Engineering, Shanghai Jiao Tong University, Shanghai, China}
% \affil[2]{Tencent, Shanghai, China}
% \affil[ ]{Emails: \{evonwang, chaixinning, rainbowow, zxcheng, xierong, song\_li\}@sjtu.edu.cn, barryjzhao@tencent.com}
\maketitle
\begin{abstract}
Recent advancements in video super-resolution (VSR) models have demonstrated impressive results in enhancing low-resolution videos. However, due to limitations in adequately controlling the generation process, achieving high fidelity alignment with the low-resolution input while maintaining temporal consistency across frames remains a significant challenge. In this work, we propose Semantic and Temporal Guided Video Super-Resolution (SeTe-VSR), a novel approach that incorporates both semantic and temporal-spatio guidance in the latent diffusion space to address these challenges. By incorporating high-level semantic information and integrating spatial and temporal information, our approach achieves a seamless balance between recovering intricate details and ensuring temporal coherence. Our method not only preserves high-reality visual content but also significantly enhances fidelity. Extensive experiments demonstrate that SeTe-VSR outperforms existing methods in terms of detail recovery and perceptual quality, highlighting its effectiveness for complex video super-resolution tasks.
\end{abstract}


\section{Introduction}
\label{sec:intro}
Video super-resolution (VSR) aims to enhance the spatial resolution of low-resolution video frames by recovering fine-grained details and improving visual quality, ultimately generating more realistic images with higher quality. Unlike traditional image super-resolution, which only focuses on enhancing individual frames, VSR seeks to maintain temporal consistency across frames, ensuring that the enhanced frames remain coherent and stable over time. 

In recent years, diffusion models\cite{ddpm} have gained considerable attention as a powerful class of generative models. The introduction of Latent Diffusion Models (LDM)\cite{stable} has significantly reduced computational demands by encoding pixel inputs to a smaller latent space, thus enhancing the applicability of diffusion models in tasks like image\cite{imagen} and video generation\cite{videocrafter2,svd}. Video super-resolution methods based on diffusion models\cite{mgldvsr,stablevsr,upscale,diffir2vr,sateco} have made notable progress, effectively addressing the blurring issues typically seen in traditional video super-resolution approaches, and are capable of generating finer, more realistic details. 

However, despite these advancements, there are still significant challenges to overcome. One of the most pressing issues lies in the misalignment between the generated outputs and the input low-resolution (LR) frames, especially under complex degradation scenarios. This challenge is further compounded by the limited ability to fully control the generation process using only the information available in the LR frames, resulting in inconsistencies and a loss of fidelity in the reconstructed outputs. 
%Such issues undermine the perceptual quality of reconstructed videos and limit their applicability in real-world settings.
Methods like MGLD-VSR\cite{mgldvsr} and StableVSR\cite{stablevsr} primarily condition on LR input frames to achieve alignment. However, their reliance on LR frames alone often fails to establish precise alignment, particularly in scenarios with complex degradations, leading to artifacts and inconsistencies between the generated outputs and input frames. Upscale-A-Video\cite{upscale} attempts to enhance visual quality by incorporating text prompts through Classifier-Free Guidance (CFG)\cite{cfg}. While this approach introduces global contextual guidance, simple text prompts are often insufficient to capture and restore intricate degraded features in complex video inputs. 
% Another challenging issue is maintaining video temporal consistency. To tackle this issue, some studies have attempted to extend image super-resolution diffusion models to the video domain. For instance, DiffIR2VR\cite{diffir2vr} proposes a training-free approach that enhances frame consistency through video token merging strategy. Other approaches\cite{mgldvsr,upscale} have incorporated temporal convolutions and temporal attention layers into the U-Net and VAE architecture to improve video stability and coherence. Additionally, some methods\cite{mgldvsr,upscale,stablevsr} integrate optical flow conditions during the sampling process to guide video generation, further enhancing temporal consistency and the preservation of motion details.
 % Thus, integrating high-level semantic features can provide the model with a deeper understanding of the scene, aiding in more accurate detail recovery and structural restoration, especially when confronted with complex degradation.

To address this, we propose a method that improves fidelity to the input frames while preserving high-reality visual quality. Misalignment with the input frames is often attributed to the excessive reliance on low-level information, such as pixels and textures, which are often heavily distorted under severe degradation. High-level semantic information, on the other hand, typically remains relatively robust, even in complex scenarios, offering a more stable and informative signal for guiding the restoration process. Previous methods\cite{seesr,xpsr,pasd} have shown that incorporating semantic guidance  is beneficial for fine-grained detail recovery. To overcome the challenges of achieving precise alignment and detail recovery, we propose the Semantic Alignment Module (SeAM). By integrating high-level semantic embeddings extracted from SAM2\cite{sam2} into the denoising U-Net, SeAM enables the model to effectively bridge the gap between the degraded input and the reconstructed output. These semantic embeddings, enriched by SAM2's zero-shot generalization capability, provide a global contextual understanding of the scene, empowering the model to restore fine details and structural integrity with greater accuracy. 
%Through this integration, SeAM addresses the shortcomings of previous methods, particularly in handling complex or unseen real-world degradations, enhancing both the alignment and robustness of the restoration process.

To further improve alignment between generated outputs and input LR frames, we introduce the Temporal-Spatio Awareness Module (TSAM). While the Semantic Alignment Module (SeAM) focuses on leveraging high-level semantic information to achieve accurate spatial alignment, ensuring consistent alignment across frames necessitates the integration of both spatial and temporal information. TSAM addresses this by facilitating the interaction and fusion of spatio-temporal features, enabling the model to capture dependencies not only within a single frame but also across adjacent frames. By harmonizing both spatial and temporal information,  TSAM enhances the model's ability to achieve precise alignment, recover fine details, and maintain smooth transitions between frames.

The primary contributions of this work are summarized as follows.
\begin{itemize}
\item We propose a novel diffusion-based video super-resolution framework that incorporates semantic and spatio-temporal understanding to effectively handle complex degradations, delivering high-quality video outputs with enhanced detail and coherence.
\item We introduce the Semantic Alignment Module (SeAM), which extracts high-level semantics from SAM2 for better detail restoration and robustness.
\item We develop the Temporal-Spatio Awareness Module (TSAM) to futher integrate spatial and temporal information, balancing fine detail recovery and cross-frame consistency.
\item Extensive quantitative and qualitative experiments demonstrate the superior performance of our proposed method, achieving state-of-the-art results in terms of both realism and fidelity.
\end{itemize}

\section{Related Works}
\label{sec:realted}
\subsection{Video Super-Resolution (VSR)}
Video super-resolution (VSR) aims to enhance the resolution of low-quality videos by utilizing both spatial and temporal information. Most CNN-based video super-resolution models\cite{basicvsr,basicvsrpp,realbasicvsr,rvrt,dbvsr} adopt lightweight architectures. To improve temporal consistency, BasicVSR\cite{basicvsr} introduces bidirectional propagation and feature alignment modules. Building upon\cite{basicvsr}, RealBasicVSR\cite{realbasicvsr} proposes data pre-cleaning module to reduce the propagation of noise and artifacts.

However, CNN-based models still struggle to generate fine-grained features. As a generative model, diffusion models have demonstrated tremendous potential in image and video generation, leading to the emergence of several diffusion-based video super-resolution algorithms\cite{mgldvsr,upscale,stablevsr,sateco,diffir2vr}. To improve video inter-frame continuity, researchers have proposed enhanced sampling strategies\cite{mgldvsr,upscale,stablevsr} and the integration of temporal modules\cite{mgldvsr,upscale,sateco} based on pre-trained diffusion models.

\subsection{Semantic Guidance in Image Super-Resolution}
In recent years, several diffusion-based image super-resolution methods\cite{seesr,xpsr,pasd} have effectively integrated semantic guidance to improve texture and structural recovery. PASD\cite{pasd} employs pre-trained high-level nets to extract high-level image information. SeeSR\cite{seesr} introduces a degradation-aware prompt extractor that generates representation embeddings and image labels, thereby enhancing the perception capabilities of diffusion T2I models. XPSR\cite{xpsr} leverages advanced multimodal large language models (MLLMs) to extract both high-level and low-level semantic embeddings, further enriching the semantic information utilized during image restoration.
\begin{figure}[htbp]
\vspace{-5mm} 
\centerline{\includegraphics[width=0.8\linewidth]{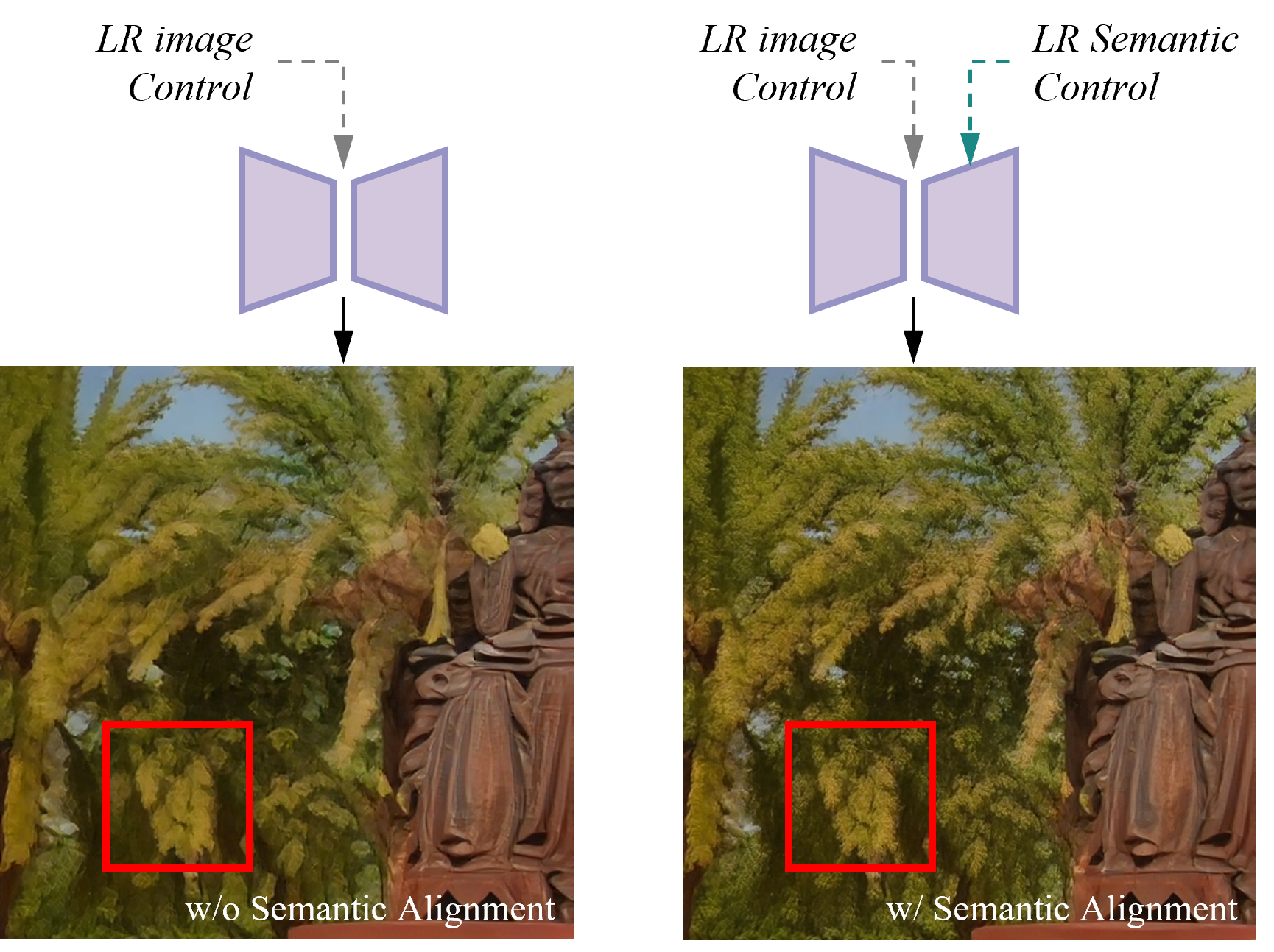}}
\caption{Comparison of VSR results with and without semantic guidance.}
\label{intro}
\vspace{-2mm} 
\end{figure}
\subsection{Temporal Consistency in Diffusion Video Generation}
Advancements in diffusion-based video generation have focused on enhancing temporal consistency. To enhance temporal coherence, TokenFlow\cite{tokenflow} employs cross-frame token propagation, while VidToMe\cite{vidtome} utilizes token merging across frames.
FLATTEN\cite{flatten} uses optical flow to compute the trajectories of image patches, thereby guiding the attention mechanism across patches.

\begin{figure*}[htbp]
\centerline{\includegraphics[width=\linewidth]{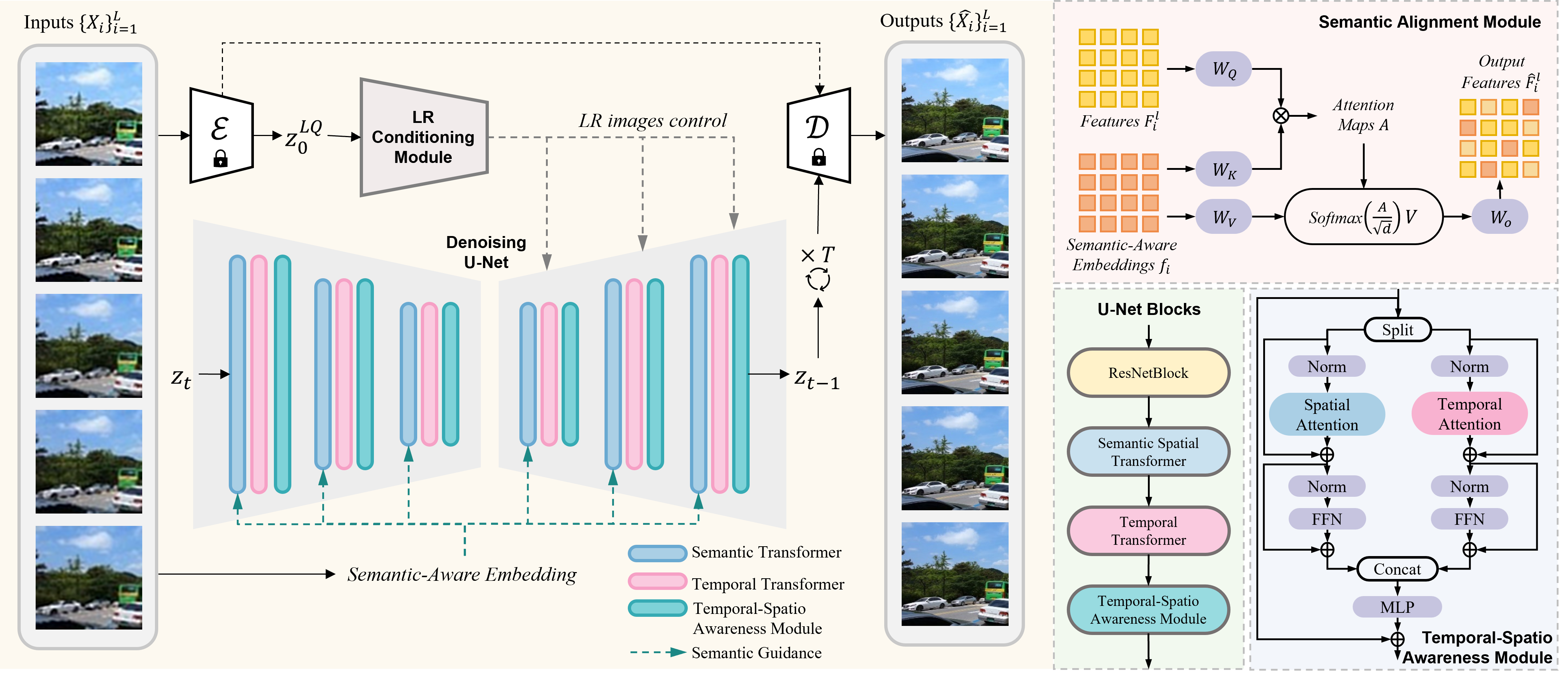}}
\caption{Overview of our proposed SeTe-VSR. SeTe-VSR enhances low-resolution video frames through two key modules. First, the Semantic Alignment Module (SeAM) leverages high-level semantic embeddings from the video frames, providing crucial scene understanding to improve detail restoration and robustness. Temporal-Spatio Awareness Module (TSAM) is employed in denoising process to integrate both spatial and temporal information, ensuring improved fine detail recovery and cross-frame consistency.}
\label{pipeline}
\end{figure*}

\section{Method}

\subsection{Overview}
 Given a set of low-quality (LQ) video frames \( X = \{X_0, X_1, \dots, X_{N-1}\} \), the objective of our VSR method is to reconstruct high-quality (HQ) video frames \( \hat{X} = \{\hat{X}_0, \hat{X}_1, \dots, \hat{X}_{N-1}\} \).

In this paper, we propose a diffusion-based video super-resolution framework that integrates both semantic and spatio-temporal understanding, allowing for the handling of complex degradations.  The overall framework of our proposed method is illustrated in Fig.~\ref{pipeline}. 

Initially, the LQ frames are divided into several segments, each containing \( L \) frames. These video segments are then passed through VAE encoder $\mathcal{E}$ to obtain the corresponding latent codes. Simultaneously, we use SAM2 to extract semantic embeddings from these video segments. The latents are noised and subsequently fed into a denoising U-Net for T steps of denoising. LR conditioning module is used to provide LR image guidance. During the denoising process, the extracted semantic embeddings are injected into the U-Net using a semantic spatial transformer, which assists in handling complex degradations, improving the model's ability to restore high-quality video frames. Additionally, Temporal-Spatio Awareness Module is incorporated into the U-Net to integrate spatial and temporal information.

During training, we optimize the denoising objective:
\begin{equation}
\mathcal{L} = \mathbb{E}_{z_0, t,c, \epsilon \sim \mathcal{N}} \left\| \epsilon - \epsilon_{\theta}(z_t; t,c) \right\|_2^2 
\label{eq:diffusion_loss}
\end{equation}

Next, we provide a description of the Semantic Alignment Module in Sec.~\ref{sub_sec:semantic}, followed by a detailed explanation of the Temporal-Spatio Awareness Module in Sec.~\ref{sub_sec:spatio_temporal} and finally training strategy in Sec.~\ref{sub_sec:training}.

\subsection{Semantic Alignment Module}
\label{sub_sec:semantic}

As shown in Fig.~\ref{pipeline}, the low-resolution (LR) video frames $\left\{X_i\right\}$ are first passed through a frozen SAM2 model to extract semantic image embeddings $\left\{f_i\right\}$, as shown in Eq.~\ref{eq:semantic_eq1}:

\begin{equation}
\label{eq:semantic_eq1}
\begin{aligned}
f_i =\verb|SAM2|(X_i) \\
\end{aligned}
\end{equation}
SAM2 extracts high-level semantic features from images, preserving crucial semantic information even under degradation. These semantic embeddings are then incorporated into the denoising U-Net via a semantic attention mechanism. Specifically,  for the $l$-th layer of U-Net,  the query vector Q is extracted from the spatial feature $F_i^l$, while the key vector K and value vector V are extracted from the semantic embedding $f_i$, as described in Eq.~\ref{eq:semantic_eq2}:
\begin{equation}
\label{eq:semantic_eq2}
\begin{aligned}
& Q = W_q(F_i^l), K = W_k(f_i),  V = W_v(f_i) \\
&\verb|Attention|(Q,K,V)=\verb|Softmax|(\frac{QK^T}{\sqrt{d}})V \\
\end{aligned}
\end{equation}
% Thus, the Semantic Alignment Module integrates high-level semantic embeddings extracted from SAM2 into the denoising U-Net, enabling the model to better understand the video scene and  handle complex visual degradations. 

As shown in Fig.~\ref{intro}, the Semantic Alignment Module leverages these high-level semantic embeddings, enabling the model to recover fine details and structural integrity with enhanced realism and fidelity, significantly enhancing its ability to handle complex visual degradations.

\subsection{Temporal-Spatio Awareness Module}
\label{sub_sec:spatio_temporal}
In video super-resolution, addressing both spatial and temporal degradation is crucial for restoring high-quality frames.
%, as damaged regions often require information from both the spatial context and adjacent frames to recover fine details. 
Relying on spatial features or temporal information alone may not be sufficient to recover fine details across frames.  To tackle these challenges, we propose the Temporal-Spatio Awareness Module, which integrates both spatial and temporal information for enhanced restoration.

The module consists of two components: a Spatial Attention Module and a Temporal Attention Module.  Specifically, for the $l$-th layer of U-Net, the output feature of temporal transformer $F^l$ is first split into spatial feature $F^l_s$ and temporal feature $F^l_t$ along channel dimension. These features are then processed separately through spatial and temporal attention mechanisms to capture spatial and temporal dependencies. The attention process for both features is shown in Eq.~\ref{eq:spatio-temporal_eq1}:
\begin{equation}
\label{eq:spatio-temporal_eq1}
\begin{aligned}
F_s^l&, F_t^l =\verb|Split|(F^l) \\
F_s^l &= \verb|SpatialAttention|( \verb|Norm|(F_s^l)) + F_s^l \\
F_s^l &= \verb|FFN|( \verb|Norm|(F_s^l)) + F_s^l \\
F_t^l &= \verb|TemporalAttention|( \verb|Norm|(F_t^l)) + F_t^l \\
F_t^l &= \verb|FFN|( \verb|Norm|(F_t^l)) + F_t^l
\end{aligned}
\end{equation}

Subsequently, the spatial and temporal features are concatenated along the channel dimension, combining both spatial and temporal information. This fused representation is then passed through a MLP layer for further integration, as shown in Eq.~\ref{eq:spatio-temporal_eq2}:
\begin{equation}
\label{eq:spatio-temporal_eq2}
\begin{aligned}
\hat{F^l}=\verb|MLP|(\verb|Concat|(F_s^l, F_t^l))+F^l
\end{aligned}
\end{equation}
\begin{table*}[htbp]
\caption{Comparison of different video super-resolution methods on various datasets. \textcolor{red}{Red} and \textcolor{blue}{blue} represent the best and second best score, respectively.}
\begin{center}
\begin{tabular}{c|c|c@{\hspace{10pt}}c@{\hspace{6pt}}c@{\hspace{6pt}}c@{\hspace{6pt}}c@{\hspace{6pt}}c@{\hspace{6pt}}c@{\hspace{6pt}}c@{\hspace{6pt}}c}
        \toprule
        Datasets & Metrics & Bicubic & DBVSR\cite{dbvsr} & BasicVSR++\cite{basicvsrpp} & RVRT\cite{rvrt}  & RealBasicVSR\cite{realbasicvsr} &  StableVSR\cite{stablevsr} & MGLD-VSR\cite{mgldvsr} & Ours \\ 
        \midrule
        \multirow{5}{*}{REDS4} 
        & PSNR $\uparrow$  &22.52&	22.60&	22.60&	22.61&	 \textcolor{red}{22.69}&	22.60&	22.59& \textcolor{blue}{22.65} \\ 
        % & SSIM $\uparrow$ &0.5710&	0.5703&	0.5702&	0.5705&	 \textcolor{red}{0.6144}&	0.5666&	0.5907& \textcolor{blue}{0.5927}\\ 
        & LPIPS $\downarrow$ &0.5360&	0.5297&	0.5303&	0.5301&	0.3411&	0.5256&	\textcolor{blue}{0.3188}&  \textcolor{red}{0.3150}\\ 
        % & NIQE $\downarrow$  &9.801&	9.852&	9.934&	9.900&	3.458&	9.147&	\textcolor{blue}{3.240}&  \textcolor{red}{3.167}\\  
        & BRISQUE $\downarrow$ & 71.16&	71.85&	71.86&	72.37&	14.96&	66.27&	\textcolor{blue}{12.66}& \textcolor{red}{10.33}\\ 
        & CLIPIQA $\uparrow$ &0.1967	&0.2067	&0.2068	&0.2082	&\textcolor{red}{0.3408}&	0.1125	&0.3343	&\textcolor{blue}{0.3368}\\
         & DOVER $\uparrow$ & 0.0263&	0.0278&	0.0282	&0.0276	&0.5095	&0.0272&	\textcolor{blue}{0.5254}	&\textcolor{red}{0.5392}\\

        % & $E^*_{warp} \downarrow$ &  \textcolor{red}{1.097}&	1.144&	1.132&	\textcolor{blue}{1.142}&	1.875&	2.056&	3.124& 2.946\\  
        \midrule
        \multirow{5}{*}{SPMCS} 
        & PSNR $\uparrow$ & 22.80&	\textcolor{blue}{22.85}&	22.75&	\textcolor{red}{22.90}&	22.72&	22.74&	22.82&	22.66\\ 
        % & SSIM $\uparrow$ & 0.5412&	0.5432&	0.5386&	0.5456&	0.5463&	0.5311&	\textcolor{red}{0.5624}&	\textcolor{blue}{0.5591}\\ 
        & LPIPS $\downarrow$ & 0.5007&	0.4940&	0.4927&	0.4916&	0.3876&0.4971	&	\textcolor{blue}{0.3589}&	\textcolor{red}{0.3532}\\ 
        & BRISQUE $\downarrow$ & 70.29&	67.93&	66.25&	67.88&	\textcolor{red}{16.16}&58.97	&	22.56&	\textcolor{blue}{20.50}\\ 
        & CLIPIQA $\uparrow$ &0.2791&	0.2834&	0.2913&	0.2947&	0.4411&0.1759	&	\textcolor{blue}{0.4491}&	\textcolor{red}{0.4813}\\
        &DOVER $\uparrow$ & 0.0440&	0.06145&	0.0621&	0.0610&	\textcolor{blue}{0.4724}& 0.0689	 &	0.4528&	\textcolor{red}{0.4774}\\
        \midrule
        \multirow{4}{*}{VideoLQ} 
        % & NIQE $\downarrow$ & 8.062&	6.720&	6.214&	6.589&	3.692&	5.845&	3.488& 3.324 \\ 
        & BRISQUE $\downarrow$ & 64.77&	61.18&	60.50&	62.18&	24.55&	47.51&	 \textcolor{blue}{22.27}& \textcolor{red}{18.58}  \\ 
        & MUSIQ $\uparrow$ & 22.55&	29.02&	28.69&	28.42&	\textcolor{blue}{55.97}&	26.86&	54.33&   \textcolor{red}{56.55}\\
        & CLIPIQA $\uparrow$ & 0.2948&	0.2700&	0.2782&	0.2813&	\textcolor{blue}{0.3918}&	0.1761&	0.3803& \textcolor{red}{0.4199} \\ 
        & DOVER $\uparrow$ & 0.3536&	0.4386&	0.4384&	0.4315 &	0.7162&	0.4338&	\textcolor{blue}{0.7252}&  \textcolor{red}{0.7431}\\
        % & $E^*_{warp} \downarrow$ & \textcolor{red}{0.6637} & 1.086 & \textcolor{blue}{0.993}	&1.009 &1.080 &1.200& 2.148& 1.815\\ 
        \bottomrule
\end{tabular}
\label{quantitative}
\end{center}
\end{table*}
By explicitly modeling the interaction between spatial and temporal features, this module enables the model to effectively leverage both information sources, leading to improved restoration quality and more accurate frame detail recovery.

% This approach enables the model to leverage both spatial and temporal information in a more structured and effective manner, rather than relying on a single type of feature. By explicitly modeling the interaction between spatial and temporal features, the Temporal-Spatio Awareness Module ensures that both types of information contribute to the restoration process, improving the handling of complex feature interactions and fine detail recovery across frames.

\subsection{Training Strategy}
Our training approach consists of two stages. In the first stage, we remove the both temporal transformer and temporal-spatio awareness module, focusing solely on the semantic spatial transformer, which helps the model learn spatial features and semantic representations independently.
In the second stage, we introduce the temporal transformer and temporal-spatio awareness module, in order to capture spatial and temporal dependencies between frames. During this stage, we freeze the other layers and optimize newly introduced modules.
\label{sub_sec:training}

\section{Experiment}
\subsection{Datasets and Implementation}
\noindent\textbf{Implementation Details.}
 The denoising U-Net is initialized using the pre-trained weights from Stable Diffusion V2.1\cite{stable}. Similar to MGLD-VSR\cite{mgldvsr}, we incorporate a VAE decoder with temporal layers. During training, we use the Adam optimizer\cite{adam} with a batch size of 3 and a constant learning rate of $1e-4$ for the first stage. In the second stage, we use the Adam optimizer\cite{adam} with a batch size of 3 and a constant learning rate of $5e-5$. During inference, we use DDPM\cite{ddpm} sampling for 50 steps for each video sequence. All experiments are implemented on a single NVIDIA A100-80G GPU.

\noindent\textbf{Training and Testing Datasets.}
For the training set, we combined the REDS\cite{reds} training and validation sets, reserving four video sequences for validation. Training sequence pairs were generated by applying the degradation pipeline from RealBasicVSR\cite{realbasicvsr}. For synthetic testing datasets, we selected REDS4\cite{reds} and SPMCS\cite{dbvsr}, containing 4 and 30 video sequences respectively. Both datasets were processed using the same degradation pipeline applied during training. For real-world testing, we utilized VideoLQ\cite{realbasicvsr}, a dataset containing 50 real-world video sequences, each exhibiting various types of degradation.

\noindent\textbf{Evaluation Metrics.}
In this study, we employ a comprehensive set of evaluation metrics to assess the performance of the proposed method. Pixel-wise accuracy is quantified using PSNR. Perceptual quality is evaluated using LPIPS\cite{lpips}. Video quality can be comprehensively evaluated through the video-specific DOVER\cite{dover} metric, which integrates technical and aesthetic dimensions. Additionally, no-reference quality metrics, such as MUSIQ\cite{musiq}, BRISQUE\cite{brisque} and CLIP-IQA\cite{clipiqa}, are utilized to evaluate the quality of real-world low-quality datasets.

\subsection{Comparisons}
We compare our proposed method with several state-of-the-art VSR methods, including DBVSR\cite{dbvsr}, BasicVSR++\cite{basicvsrpp}, RVRT\cite{rvrt}, RealBasicVSR\cite{realbasicvsr}, as well as diffuison-based methods StableVSR\cite{stablevsr} and MGLD-VSR\cite{mgldvsr}.

\noindent\textbf{Quantitative Comparison.}
As shown in Table~\ref{quantitative}, our approach achieves the highest LPIPS across all synthetic datasets, indicating its superior perceptual quality. While PSNR is a widely used metric for evaluating VSR tasks, it primarily focuses on pixel-wise accuracy and often fail to capture perceptual aspects, such as texture realism and structural coherence. When evaluated on real-world VSR datasets, our method further excels by securing the highest BRISQUE, MUSIQ and CLIP-IQA scores, highlighting its proficiency in generating realistic textures and fine-grained details. Additionally, our approach ranks first in DOVER, emphasizing its capacity to produce content with high visual consistency and perceptual quality.
\begin{figure*}[htbp]
    \centering
    % 左边单张图
    \begin{minipage}{0.2183\textwidth}
        \centering
        \captionsetup{labelformat=empty, font=footnotesize}
        \includegraphics[width=\textwidth]{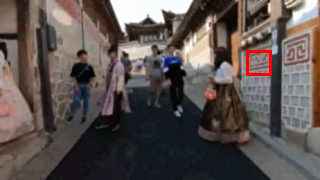} % 修改为图片
        \caption*{Sequence 011 of REDS4}
    \end{minipage} \hfill
    % 右边四列每列两张图
    \begin{minipage}{0.77\textwidth}
        \centering
            \begin{minipage}{0.16\linewidth}
            \centering
            \captionsetup{labelformat=empty, font=footnotesize} % 只对这个子图应用 labelformat=empty
            \includegraphics[width=\linewidth]{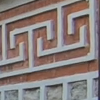}
            \caption*{GT} % 使用 \caption* 避免编号
            \label{GT}%文中引用该图片代号
        \end{minipage}
            \begin{minipage}{0.16\linewidth}
            \centering
            \captionsetup{labelformat=empty, font=footnotesize} % 只对这个子图应用 labelformat=empty
            \includegraphics[width=\textwidth]{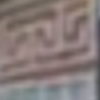}
            \caption*{Bicubic} % 使用 \caption* 避免编号
            \label{bicubic}
        \end{minipage}
        % 子图2
                \begin{minipage}{0.16\linewidth}
            \centering
            \captionsetup{labelformat=empty, font=footnotesize} % 只对这个子图应用 labelformat=empty
            \includegraphics[width=\linewidth]{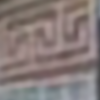}
            \caption*{RVRT\cite{rvrt}} % 使用 \caption* 避免编号
            \label{RVRT}%文中引用该图片代号
        \end{minipage}
        % 子图3
        \begin{minipage}{0.16\linewidth}
            \centering
            \captionsetup{labelformat=empty, font=footnotesize} % 只对这个子图应用 labelformat=empty
            \includegraphics[width=\linewidth]{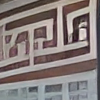}
            \caption*{RealBasicVSR\cite{realbasicvsr}} % 使用 \caption* 避免编号
            \label{RealBasicVSR}
        \end{minipage}
        % 子图4
        \begin{minipage}{0.16\linewidth}
            \centering
            \captionsetup{labelformat=empty, font=footnotesize} % 只对这个子图应用 labelformat=empty
            \includegraphics[width=\linewidth]{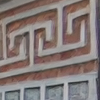}
            \caption*{MGLD-VSR\cite{mgldvsr}} % 使用 \caption* 避免编号
            \label{MGLD-VSR}%文中引用该图片代号
        \end{minipage}
        \begin{minipage}{0.16\linewidth}
            \centering
            \captionsetup{labelformat=empty, font=footnotesize} % 只对这个子图应用 labelformat=empty
            \includegraphics[width=\linewidth]{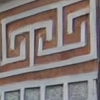}
            \caption*{Ours} % 使用 \caption* 避免编号
            \label{ours}%文中引用该图片代号
        \end{minipage}

    \end{minipage}
\caption{Qualitative comparisons of 4$\times$ video super-resolution on synthetic datasets.}
\label{reds_data}
\end{figure*}
% videolq figure1
\begin{figure*}[htbp]
    \centering
    % 左边单张图
    \begin{minipage}{0.282\textwidth}
        \centering
        \captionsetup{labelformat=empty, font=footnotesize}
        % \vspace{-0.97cm}
        \includegraphics[width=\textwidth]{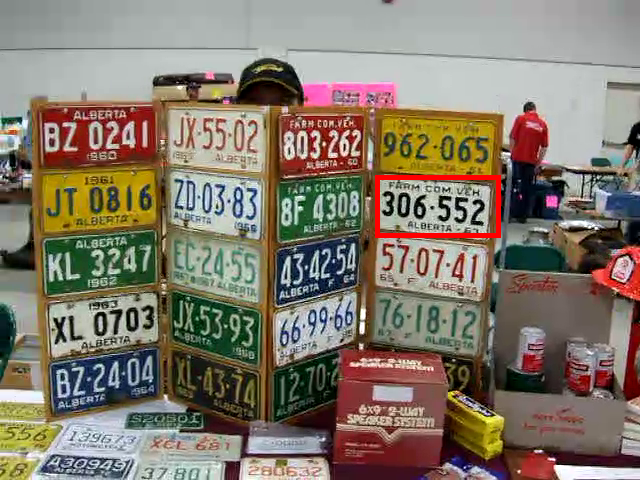} % 修改为图片
        \caption*{Sequence 041 of VideoLQ}
        % \vspace{0.23cm}
        \vspace{0.25cm}
        \includegraphics[width=\textwidth]{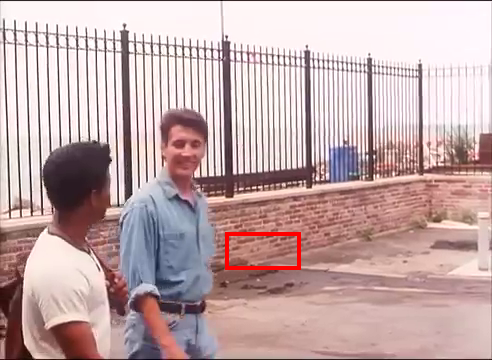} % 修改为图片
        \caption*{Sequence 004 of VideoLQ}
    \end{minipage} \hfill
    % 右边四列每列两张图
    \begin{minipage}{0.71\textwidth}
        \centering
        \captionsetup{labelformat=empty, font=footnotesize}
        \begin{minipage}{0.24\linewidth}
            \centering
            \includegraphics[width=\textwidth]{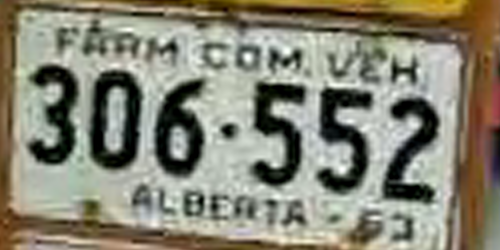}
            \caption*{Bicubic}
            \label{bicubic}
        \end{minipage}
        \vspace{0.2cm}
        \begin{minipage}{0.24\linewidth}
            \centering
            \includegraphics[width=\linewidth]{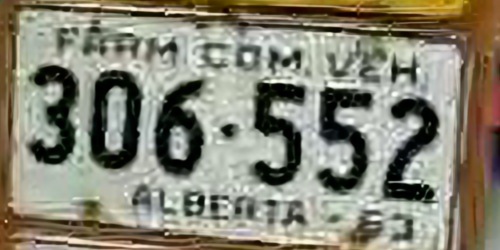}
            \caption*{BasicVSR++\cite{basicvsrpp}}
            \label{BasicVSR++}
        \end{minipage}
        \begin{minipage}{0.24\linewidth}
            \centering
            \includegraphics[width=\linewidth]{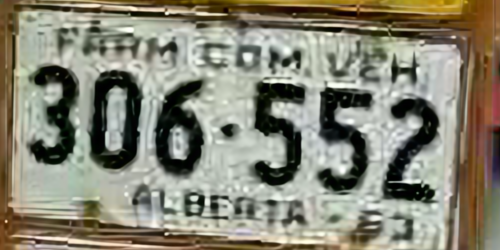}
            \caption*{DBVSR\cite{dbvsr}}
            \label{DBVSR}%文中引用该图片代号
        \end{minipage}
        \begin{minipage}{0.24\linewidth}
            \centering
            \includegraphics[width=\linewidth]{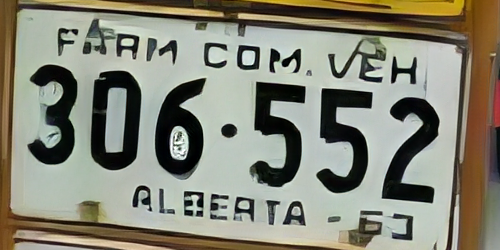}
            \caption*{RealBasicVSR\cite{realbasicvsr}}
            \label{RealBasicVSR}%文中引用该图片代号
        \end{minipage}

        \begin{minipage}{0.24\linewidth}
            \centering
            \includegraphics[width=\textwidth]{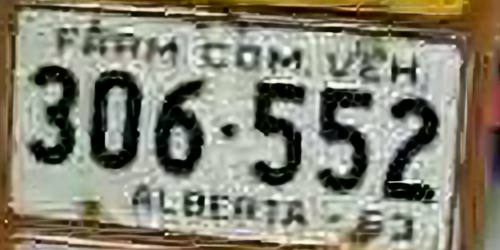}
            \caption*{RVRT\cite{rvrt}}
            \label{RVRT}
        \end{minipage}
        \vspace{0.2cm}
        \begin{minipage}{0.24\linewidth}
            \centering
            \includegraphics[width=\linewidth]{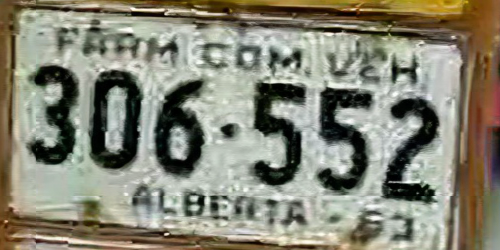}
            \caption*{StableVSR\cite{stablevsr}}
            \label{StableVSR}
        \end{minipage}
        \begin{minipage}{0.24\linewidth}
            \centering
            \includegraphics[width=\linewidth]{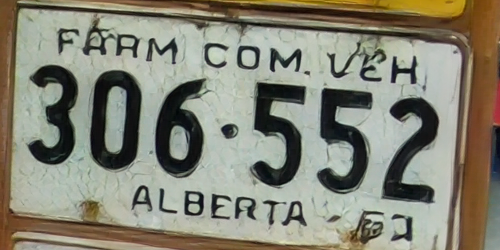}
            \caption*{MGLD-VSR\cite{mgldvsr}}
            \label{MGLD-VSR}%文中引用该图片代号
        \end{minipage}
        \begin{minipage}{0.24\linewidth}
            \centering
            \includegraphics[width=\linewidth]{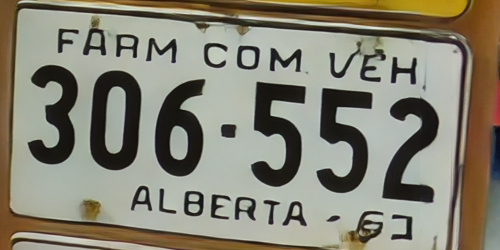}
            \caption*{Ours}
            \label{Ours}%文中引用该图片代号
        \end{minipage}

        \centering
        \begin{minipage}{0.24\linewidth}
            \centering
            \includegraphics[width=\textwidth]{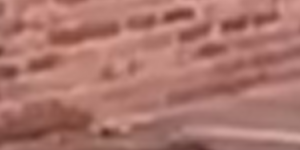}
            \caption*{Bicubic}
            \label{bicubic}
        \end{minipage}
        \vspace{0.2cm}
        \begin{minipage}{0.24\linewidth}
            \centering
            \includegraphics[width=\linewidth]{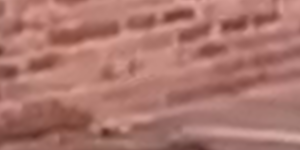}
            \caption*{BasicVSR++\cite{basicvsrpp}}
            \label{BasicVSR++}
        \end{minipage}
        \begin{minipage}{0.24\linewidth}
            \centering
            \includegraphics[width=\linewidth]{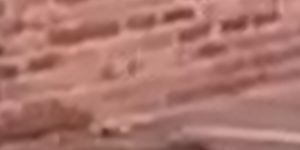}
            \caption*{DBVSR\cite{dbvsr}}
            \label{DBVSR}%文中引用该图片代号
        \end{minipage}
        \begin{minipage}{0.24\linewidth}
            \centering
            \includegraphics[width=\linewidth]{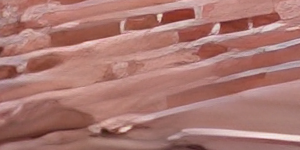}
            \caption*{RealBasicVSR\cite{realbasicvsr}}
            \label{RealBasicVSR}%文中引用该图片代号
        \end{minipage}

               \begin{minipage}{0.24\linewidth}
            \centering
            \includegraphics[width=\textwidth]{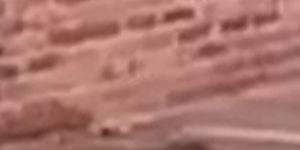}
            \caption*{RVRT\cite{rvrt}}
            \label{RVRT}
        \end{minipage}
        % \vspace{0.2cm}
        \begin{minipage}{0.24\linewidth}
            \centering
            \includegraphics[width=\linewidth]{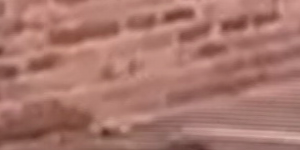}
            \caption*{StableVSR\cite{stablevsr}}
            \label{StableVSR}
        \end{minipage}
        \begin{minipage}{0.24\linewidth}
            \centering
            \includegraphics[width=\linewidth]{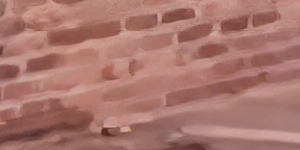}
            \caption*{MGLD-VSR\cite{mgldvsr}}
            \label{MGLD-VSR}%文中引用该图片代号
        \end{minipage}
        \begin{minipage}{0.24\linewidth}
            \centering
            \includegraphics[width=\linewidth]{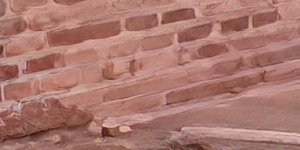}
            \caption*{Ours}
            \label{Ours}%文中引用该图片代号
        \end{minipage}
    \end{minipage}
\caption{Qualitative comparisons of 4$\times$ video super-resolution on real-world dataset.}
\label{real_data}
\vspace{-0.1cm}
\end{figure*}
\begin{figure}[htbp]
    \centering
     \begin{minipage}{0.13\textwidth}
        \centering
        \includegraphics[width=\linewidth]{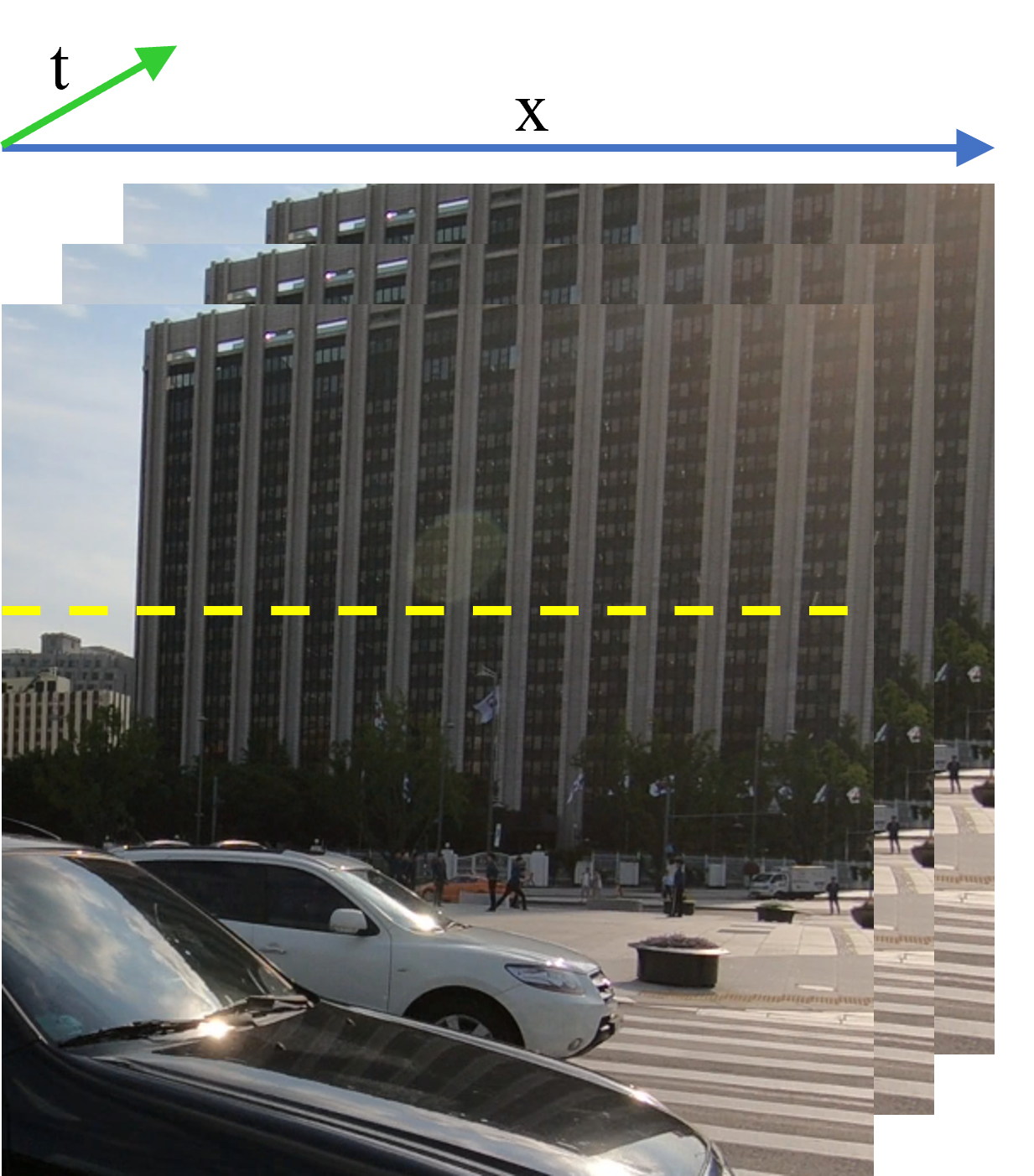}   
    \end{minipage}%
    \begin{minipage}{0.27\textwidth}
        \centering        
        \includegraphics[width=\linewidth]{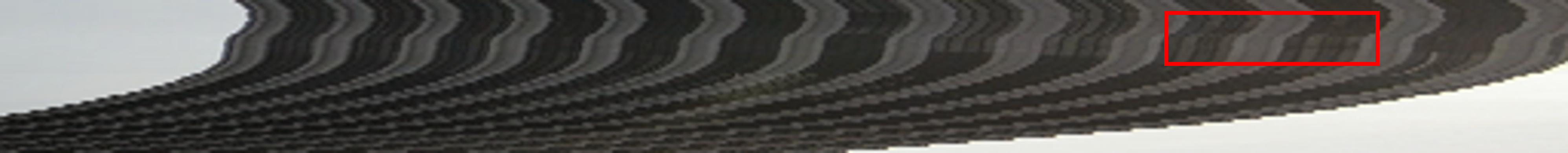}
        
        \vspace{0.05cm}  % 增加间隔
                
        \includegraphics[width=\linewidth]{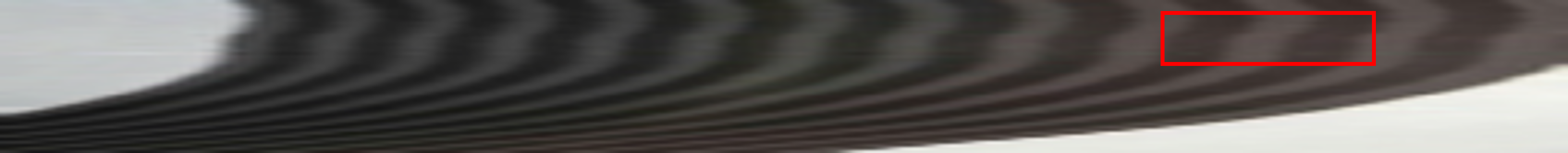}  

        \vspace{0.05cm}

        \includegraphics[width=\linewidth]{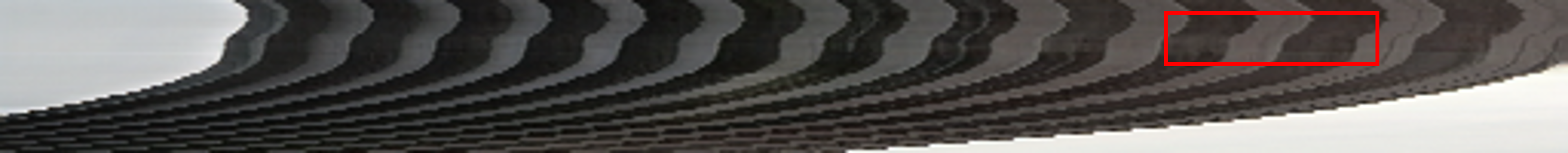}  
        
         \vspace{0.05cm}

        \includegraphics[width=\linewidth]{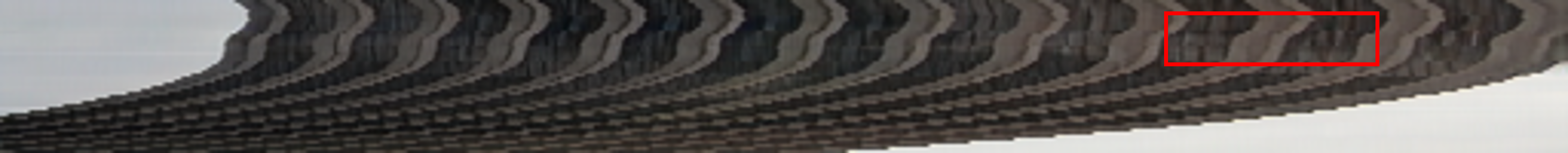}  
        
        \vspace{0.05cm}  % 增加间隔
        
        \includegraphics[width=\linewidth]{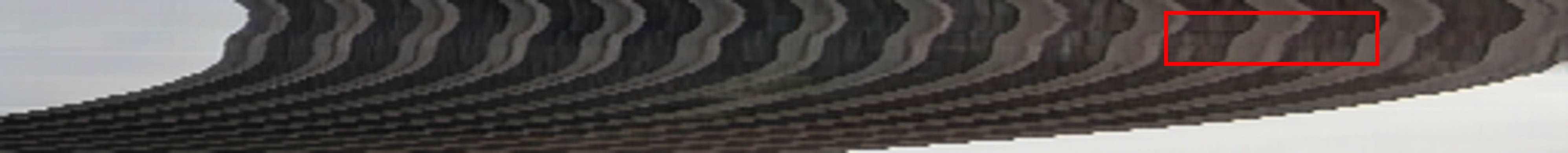}  
 
    \end{minipage}%
 \hspace{0.03cm}
    \begin{minipage}{0.08\textwidth}
        \raggedright
        {\scriptsize
        GT \\
         \vspace{0.25cm}  % 增加间隔
        Bicubic \\
        \vspace{0.25cm}  % 增加间隔
        RealBasicVSR \\
        \vspace{0.25cm}  % 增加间隔
        MGLD-VSR \\
        \vspace{0.25cm} 
        Ours \\
        }
    \end{minipage}
    \caption{Temporal profile comparison of different VSR methods.}
    \label{temporalprofile}
    \vspace{-0.5cm}
\end{figure}

\noindent\textbf{Qualitative Comparison.}
We present the visual results of the proposed method in Fig.~\ref{reds_data} and Fig.~\ref{real_data}, which show performance on the synthetic and real-world datasets, respectively. As shown in Fig.~\ref{reds_data}, on the synthetic dataset, our method demonstrates superior fidelity by accurately recovering fine textures and sharp edges while preserving structural integrity. In contrast, RVRT produces blurry outputs, failing to capture intricate details, while RealBasicVSR and MGLD-VSR introduce various distortions that compromise visual authenticity. As shown in  Fig.~\ref{real_data}, on the real-world dataset, our method effectively mitigates real-world degradations and restores fine details, yielding high-quality results with improved clarity and realism. In comparison, other methods either fail to fully eliminate the degradations or generate unrealistic artifacts, ultimately reducing the overall visual quality of the restored frames.

\noindent\textbf{Temporal Consistency.}
Our approach is designed to achieve a balance between maintaining smooth temporal transitions and reconstructing high-quality details. The temporal profile, illustrated in Fig.~\ref{temporalprofile}, provides a comparative analysis of the temporal consistency between our proposed method and other approaches. It demonstrates that our method not only preserves the intricate details within each frame but also ensures these details transition smoothly over time. 

% Since blurred images typically lack fine details and edge information, the inter-frame variations tend to be smoother, leading to a lower $E^{*}_{warp}$ in blurry videos. Our approach, on the other hand, prioritizes fine-grained detail restoration, particularly in areas with complex textures.
% This focus introduces more complex motion dynamics and greater inter-frame variations, resulting in a higher $E^{*}_{warp}$ compared to approaches that prioritize smoother temporal transitions over detailed recovery. Additionally, the temporal profile, illustrated in Fig.~\ref{temporalprofile}, provides a comparative analysis of the temporal consistency between our proposed method and other approaches. Compared to other methods, which may generate overly smooth results, our approach focuses on preserving intricate details, while maintaining temporal consistency.

\subsection{Ablation Study}
To thoroughly evaluate the effectiveness of each component in our proposed SeTe-VSR, we conduct an ablation study on the REDS4 dataset, with the experimental results presented in Table~\ref{ablation}. The results demonstrate that the integration of semantic guidance leads to significant improvements in perceptual quality. Additionally, the temporal-spatial awareness module enhances both temporal consistency and overall video quality. When combined, the full model strikes an optimal balance between generation quality and temporal consistency.

\begin{table}[htbp]
\caption{Quantitative comparison of ablation studies.}
\begin{center}
% \begin{tabular}{|p{0.8cm}|p{1cm}p{1cm}|p{1cm}p{1cm}p{1.2cm}|}
\begin{tabular}{|c|cc|c@{\hspace{6pt}}c@{\hspace{6pt}}c|}
\hline
\textbf{Exp.} & SeAM & TSAM & LPIPS $\downarrow$ &  BRISQUE  $\downarrow$  &   DOVER  $\uparrow$  \\
\hline
(a) &  &  & 0.3351 &	13.60&	0.4800\\
(b) & \checkmark & & \textbf{0.3147}&	\underline{10.70}&	\underline{0.5280}\\
(c) & &\checkmark & 0.3398&	14.04&		0.4899  \\
(d) &\checkmark & \checkmark& \underline{0.3150}&	\textbf{10.33}&	\textbf{0.5392}\\
\hline
\end{tabular}
\label{ablation}
\end{center}
\end{table}

\section{Conclusion}
In this paper, we proposed SeTe-VSR, a novel approach that leverages semantic and temporal guidance within the latent diffusion framework to address the challenges in real-world video super-resolution (VSR). By introducing the Semantic Alignment Module (SeAM), we enhanced fine-grained detail restoration through high-level semantic embeddings, improving robustness and generalization across diverse degradation scenarios. Additionally, the Temporal-Spatio Awareness Module (TSAM) facilitated effective integration of spatial and temporal information, ensuring both fine detail recovery and temporal consistency. Our extensive experiments demonstrate that SeTe-VSR outperforms existing methods, achieving state-of-the-art performance in visual quality and temporal coherence. 
% \section*{Acknowledgment}

% The preferred spelling of the word ``acknowledgment'' in America is without 
% an ``e'' after the ``g''. Avoid the stilted expression ``one of us (R. B. 
% G.) thanks $\ldots$''. Instead, try ``R. B. G. thanks$\ldots$''. Put sponsor 
% acknowledgments in the unnumbered footnote on the first page.
\section{Acknowledgment}
This work was partly supported by the Fundamental Research Funds for the Central Universities, STCSM under Grant 22DZ2229005, 111 project BP0719010.
\bibliographystyle{IEEEbib}
% \bibliography{icme2025references}
\bibliography{icme2025_template_anonymized}

\end{document}